\documentclass[smallextended]{svjour3} 
\usepackage{cite,url,comment,multirow}
\usepackage{amsmath,amssymb,amsfonts}
\usepackage{algorithm,algorithmic,enumitem,siunitx}
\usepackage{graphicx}
\usepackage{textcomp}
\usepackage{xcolor}
\graphicspath{{figs/}}

\begin{document}
\title{Correlated Initialization for Correlated Data}
\author{Johannes Schneider}
\institute{Institute of Information Systems, University of Liechtenstein, Liechtenstein \email{johannes.schneider@uni.li}  }

\date{Received: date / Accepted: date}

\maketitle



\begin{abstract}
Spatial data exhibits the property that nearby points are correlated. This also holds for learnt representations across layers, but not for commonly used weight initialization methods. Our theoretical analysis quantifies the learning behavior of weights of a single spatial filter. It is thus in contrast to a large body of work that discusses statistical properties of weights. It shows that uncorrelated initialization (i) might lead to poor convergence behavior and (ii) training of (some) parameters is likely subject to slow convergence. Empirical analysis shows that these findings for a single spatial filter extend to networks with many spatial filters. The impact of (correlated) initialization depends strongly on learning rates and l2-regularization.
\end{abstract}

\medskip


\section{Introduction} \label{sec:intro}
\begin{figure}
\centering{\includegraphics[width=0.75\linewidth,trim=4 6 4 4,clip]{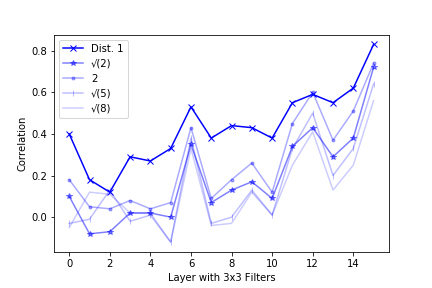}}

\caption{Distance-dependent correlation of weights of 3 $\times$ 3 filters for ResNet50 trained on ImageNet}\label{fig:cor} 
\end{figure}

Highly non-linear models such as deep learning networks are commonly subject to iterative numerical optimization, often sensitive to initialization. Ideally, weights of a network are initialized at the global optimum, rendering any form of optimization unnecessary. In the worst case, initialization is such that iterative optimization schemes either do not converge or give inferior solutions. However, deep learning networks, in general, are difficult to understand \cite{mesk20}. Thus, an important question in deep learning is \emph{How to initialize weights of a network?} This question has led to multiple works\cite{glo10,mis15,sax13,zha19,gro19}. These papers commonly draw motivation by investigating behavior across layers to foster gradient flow based on aggregate properties of weights of a layer. In contrast, this paper derives across-layer behavior from focusing on a single spatial filter, often consisting of 3 $\times$ 3 weights in modern convolutional neural networks (CNNs). 
We discuss the behavior of individual weights within a spatial filter. CNNs encode the assumption that nearby (input) locations are more relevant to learn representations than distant ones. This is justified largely by the observation that inputs are spatially correlated, meaning that there are fewer dependencies between distant than between nearby locations. However, learnt representations might show strong differences between nearby locations. For example, they might represent Gabor filters \cite{coa11} for edge detection. Still, overall correlation between adjacent filter weights is clearly positive, as indicated in Figure \ref{fig:cor}. Correlation decreases with distance and increases with layers. Similar results were obtained for additional architectures such as VGG-16, InceptionV3, Xception, and DenseNet121 (available in Keras, trained on ImageNet).\\
We provide theoretical insights on convergence behavior showing that non-correlated initialization can drastically reduce gradient flow across layers and might require longer training until convergence compared to correlated initialization in the absence of batch normalization. We also discuss common regularization schemes with respect to their impact on spatial correlation showing that L2-regularization fosters spatial correlation.  The developed theory suggests that spatial weights of $k \times k$  filters should also initially be spatially correlated. We propose multiple methods for correlated initialization that yield large improvements of several percent on accuracy for CNNs in the absence of L2-regularization and low learning rates and smaller, but still significant improvements with L2-regularization in place and large learning rates.\\ 
The paper is organized as follows: We begin with discussing related work (Section \ref{sec:relw}) and a theoretical investigation of a single spatial filter in Section \ref{sec:single}, which we extend to multiple layers in Section \ref{sec:multi}. The theoretical underpinnings inform our correlated initialization algorithm developed in Section \ref{sec:corr}, which we evaluate empirically in Section \ref{sec:eva}. Finally, we reflect on our work in Section \ref{sec:dis} and state our conclusions in Section \ref{sec:conc}.

\begin{figure*}
		\centering{
			\includegraphics[width=\linewidth]{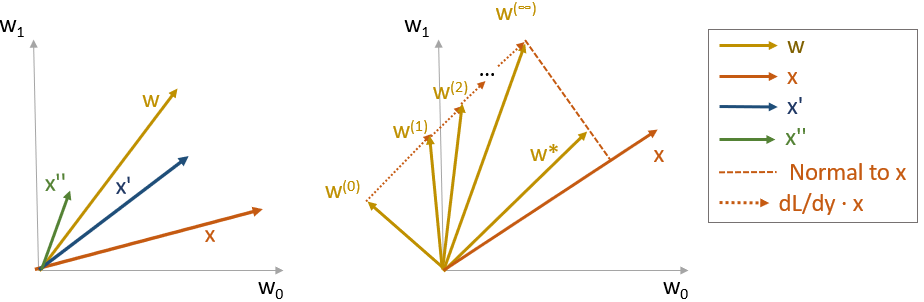}}
		\caption{Left: Setup: 3 training samples $X$, $X'$ and $X''$ and spatial filter $w=(w_0,w_1)$. Right: Training for one sample starting at $w^{(0)}$. The final solution $w^{\infty}$ is far from the optimal $w^*$.} \label{fig:setup} 
\end{figure*}
    
    \begin{figure*}
			\centering{
			\includegraphics[width=\linewidth]{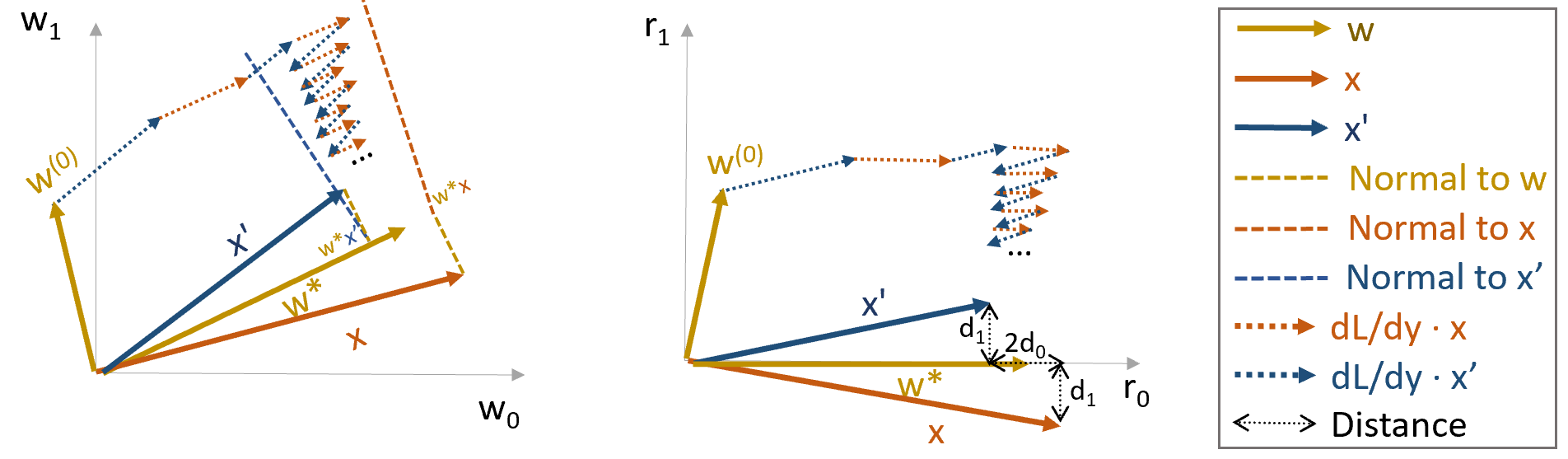}}
		\caption{Training of two weights of a spatial filter for two training samples. Right: Rotated coordinate system with distances} \label{fig:wmul}
\end{figure*}

\section{Related Work} \label{sec:relw}
Initialization schemes have often been designed to allow training of very deep networks. \cite{glo10} monitored activations and gradients and utilized the idea that variance of layer (inputs) should stay the same across layers to encourage flow of information through the network. The derivation in \cite{glo10} does not account for non-linearities. Formally, using Glorot a single parameter $w_{i,j}$ is initialized using the normal distribution: $$w_{i,j}  \sim N(0,\sqrt{\frac{2}{f_{in}+f_{out}}})$$ where $f_{in}$ and $f_{out}$ denotes the fan-in and fan-out of the layer the weight $w_{i,j}$ resides in.  A uniform distribution may also be used (adjusting the normalization constant from 2 to 6).

It has been criticized for neglecting non-linearities \cite{kum17w}. Altering the derivation including rectified linear units (ReLU) led to the Kaiming (also known as ``He'' ) initialization  \cite{he15}. Kaiming initialization seems also to be favorably related to over-parameterization of networks \cite{arp19}. Kaiming initialization using a normal distribution is given by:
$$w_{i,j}  \sim N(0,\sqrt{\frac{2}{f_{in}}})$$

Similar to these early works, more recent work has also considered activation scale\cite{han18} and gradients\cite{bal17} as well as dynamical isometry properties\cite{sax13,yan17}. In contrast, we are not primarily concerned with depth, e.g., the vanishing and exploding gradient problem, but focus on deriving initialization for a single spatial filter.\\
Initializing networks by training on another dataset is common in transfer learning or by directly extracting information from data and activation maps \cite{kotur17}. It has been said to yield better performance and faster training. Recent work\cite{he19} has challenged this wisdom by providing examples where random initialization leads to as good performance as pre-training and fine-tuning, albeit at a potential longer training time. This is aligned with our findings since we also observe larger gain in accuracy for shorter training times. Using some form of data-dependent initialization can also improve convergence speed \cite{agu19,he19}. \\ 
There is also a relationship between batch normalization\cite{iof15} and weight initialization. \cite{mis15} extends an orthonormal initialization \cite{sax13} with unit-variance. They perform batch normalization only on the very first mini-batch given orthonormal initialization. This allows them to avoid the computational costs of batch-normalization, but they do not claim any accuracy gains compared to relying on batch-normalization only. \cite{zha19} shows that using proper initialization instead of normalization (layers) for residual networks yields equal performance. The focus is primarily on avoiding issues in training due to exploding or vanishing gradients. 
Theoretical underpinnings for batch normalization are still subject to discussion. Its success has been attributed to internal covariate shift\cite{iof15}, and smoother loss surface \cite{san18}. \cite{luo18} claims that regularization in the context of batch-normalization is accomplished due to noise injection on inputs.\\
Furthermore, input-output correlations can also be analyzed based on matrix calculations, e.g., using eigenvectors \cite{adv20}. However, overall this analysis often focuses on large-scale statistical properties. In contrast, we investigate more tangible cases focusing separately on layer to layer dynamics and behavior of weights within a spatial filter. We believe that both approaches are valuable, and we view our approach as adding to the toolbox of theoreticians. 
\cite{sch19loc} derived a regularization scheme arguing that central weights in a spatial filter should be regularized less. They briefly mention initializing weights nearer to the center of a spatial filter with larger values, which is also a form of correlated initialization.
There have been extensive discussions on how information flows from the input through the network and how it is transformed \cite{sax19}. Correlated initialization facilitates the flow of information through the network from input to output. 
Furthermore, the impact of correlated samples, i.e., samples within a batch are correlated, has also been investigated for Gaussian processs \cite{chen20}. We focus more on the correlation of attributes within a sample. 
The impact of initialization scale has also been investigated \cite{meh20} showing interesting effects such as memorization of training data for certain activation functions such as ``sine''. Scaling is not a key concern in our paper, i.e., we perform standard scaling using fan-In as in \cite{he15}.

\section{Within Layer, Single Filter Convergence} \label{sec:single}
We discuss the interdependence of training data and gradient descent for a single $k \times k$ spatial filter $w$ with ReLU activation and quadratic loss function, i.e., $\hat{y}:=\max(w\cdot X,0)$ and $L(X,w):=(y-\hat{y})^2$. The optimal weights are denoted as $w^*$. We use $y(X)=w^*X$. This essentially expresses the assumption that the spatial $k \times k$ filter followed by a ReLU is a suitable model.  We discuss two scenarios: (i) The number of training samples is less than the number of parameters (under-determined system) and (ii) the number of training samples is at least as large as the number of parameters. 

\subsection{Under-determined system} Gradient descent can only explore the parameter space that is given by a linear combination of the base vectors (= span) of the training vectors with the initial vector added to it as a bias. The initialization expresses the prior belief of where a good solution is likely to be found. This belief also restricts the possible solutions that can be found. Thus, initialization has a profound impact on the quality of the solution. This can be seen by looking at the possible directions that a weight vector can be changed using gradient descent. The derivative of an aggregation layer $w\cdot x$ with respect to weight $w_i$ is given by the input, i.e., $dy/dw_i = x_i$. The derivative of the loss with respect to weight $w_i$ is: $dL/dw_i=dL/dy\cdot dy/dw_i = dL/dy\cdot x_i$. Thus, $dL/dy$ provides the magnitude of change and the inputs $x_i$ provide the direction of the gradient vector $\nabla L_w$. The weights $w_i$ do not impact direction at all and the possible directions are entirely given by $\nabla L_w$ that depends only on the training samples.\\
The convergence behavior of an under-determined system is illustrated in the left panel of Figure \ref{fig:setup} for a single training sample $(X,y)=(X,w^*X)$ and two weight vectors.\footnote{We might also assume a distorted $w^*X$, i.e., $w^*X+ \epsilon$, as discussed in more detail later.} Note that axis for weight dimensions $w_j$ and inputs $X_j$ overlap. Since there is only one training example the change of $w$ can only be in the direction of the sample $X$. Thus, starting from an initial vector $w^{(0)}$ the vector $w$ converges towards $w^{(\infty)}$ following a straight line. The final weights $w^{(\infty)}$ achieve zero training error, i.e., $\hat{y}=w^{(\infty)}\cdot X = w^{*}\cdot X=y$. Still, they are dissimilar from the optimal weights $w^*$. Therefore, performance on unseen examples is likely non-satisfactory.

\subsection{(Over)-determined system} 
If the training samples span the entire parameter space then convergence of $w$ to the optimal $w^*$ is ensured for (adequate learning rates) due to the convex nature of the problem. However, the number of iterations to get close to the optimal solution is dependent on the initial conditions. This is indicated for two training samples and two weight vectors in Figure \ref{fig:wmul}. In the right panel, each coordinate $w_j$ moves towards the optimal coordinate $w^*_j$ in either a ``zig-zag'' or ``straight'' manner. (Zig-zagging is likely to happen, if points are not evenly distributed around $w^*$.)  Both coordinates start out moving straight. Once the weight vector $w$ has crossed the normal towards $X'$, the zig-zagging begins. As shown in Figure \ref{fig:wmul}, this indicates slow convergence. To keep it to a minimum, ideally, the initial vector $w^{(0)}$ and $w^*$ point in the same direction, and samples $X$ and $X'$ are at a large angle to $w^*$. Therefore, given that weights $w^*$ are correlated, correlated initialization provides faster convergence.

Let us formalize and analyze a generalized scenario of the right panel in Figure \ref{fig:wmul}. For the optimal solution holds without loss of generality (wlog): $w^*=(w^*_0,0,\ldots,0)=(1,0,\ldots,0)$. 
The following technical assumptions are helpful to derive a closed form expression. We shall discuss loosening them after proving Theorem \ref{thm:rec}.\\

\textbf{Assumptions}:
\begin{enumerate}
\item \label{ass:pos} We assume that samples $X \in S$ have positive coordinates. This assumption is fulfilled for the prevalent CNN architectures using batch-normalization followed by a ReLU unit.
    \item \label{ass:sym} We assume a $p\geq 2$ dimensional space with $2^p$ training samples $S=\{(X,y)\}$ that are symmetric around $w^*$, i.e., $X=(1+ a_0,a_1,a_2,\ldots,a_{p-1})$, where $a_i \in \{-d_i,d_i\}$ for constants $d_i \in [0,1]$. A distance $d_i$ quantifies how much coordinate $i$ of a sample $X$ deviates from the optimal representation $w^*$. Small distances $d_i$ imply strong correlation between inputs, since in this case all samples are similar to $w^*$.
\item \label{ass:opt} We (still) assume $y(X)=w^*X$. 
\item \label{ass:act} We assume that all points are active, i.e., a point $X$ is active, if $w\cdot X>0$ for all samples $X \in S$. An active point contributes to the gradient.
\end{enumerate}

We consider gradient descent on samples $S$ for a fixed learning rate $\lambda$ that is sufficiently small to ensure convergence:
{
\begin{align} 
w^{(i+1)}&=w^{(i)}-\frac{\lambda}{|S|} \sum_{X \in S}  \nabla_w f(X,w) \label{eq:grad} \\ 
    \nabla_w f(X,w)&= \label{eq:nab}
\begin{cases}
    -2((w^*-w)\cdot X)\cdot X,& \text{if } w\cdot X>0\\
    0,              & \text{otherwise}
\end{cases}  
\end{align} }%

After proving the following theorem we explain its implications and how to loosen its assumptions:

\begin{theorem} \label{thm:rec}
For weight updates during backpropagation holds:
{
\begin{align}  
w^{(k+1)}_{0}=&w^{(k)}_{0}+2\lambda (1+d_0^2)\cdot(w^*_0-w^{(k)}_{0}) \label{eq:rec1}\\
w^{(k+1)}_{i}=&w^{(k)}_{i} -  2\lambda\big(2(w^*_0-w_{0})d_i+ w_i(d_i)^2\big) \cdot w^{(k)}_{i} \text{ for } i>0 \label{eq:rec2}
\end{align} }%

\end{theorem}

\begin{proof}
Using $w^*=(w^*_0,0,\ldots,0)$ in Equation (\ref{eq:nab}) yields the following weight updates $\nabla_w f(X,w)$:

{
\begin{align} 
-\nabla_{w_{j}} f(X,w) &= 2 \big((w^*_0-w_{0}) X_{0}- \sum_{i>0}w_{i}X_{i}\big) X_{j} \label{eq:up}
\end{align} }%

Next, we evaluate Equations (\ref{eq:grad}) and (\ref{eq:up}) for samples $S$. We consider the following subsets of $S$: All points $X \in S_{+,i} \subset S$ with the coordinate $a_i$ having non-negative sign for $d_i$, i.e., for $i=0$ we have $a_0=1+d_0$ and for $i>0$, we have $a_i=d_i$ as well as point set $S\setminus S_{+,i}$, where the sign is negative.

We use Equation (\ref{eq:up}) and investigate $w_0$ first:

{
\begin{align}
-\sum_{X \in S} \nabla_{w_{0}} f(X,w) 
&= -\sum_{X \in S_{+,0}} \nabla_{w_{0}} f(X,w)-\sum_{X \in S\setminus S_{+,0}} \nabla_{w_{0}} f(X,w) \nonumber\\
& =2\sum_{X \in S_{+,0}} \big((w^*_0-w_{0}) (1+d_0)- \sum_{i>0}w_{i}a_{i}\big) \cdot(1+d_0) \nonumber \\
& +2\sum_{X \in S\setminus S_{+,0}}\big((w^*_0-w_{0}) (1-d_0)- \sum_{i>0}w_{i}a_{i}\big) \cdot(1-d_0) \label{eq:sym}
\end{align} }%

By definition of $S$ and $S_{+,i}$ there exists for each point $X=(1+a_0,a_1,\ldots,a_{i},\ldots,a_{d-1}) \in S_{+,j}$ with $i\neq j$  a point $X'=(1+a_0,a_1,\ldots,-a_{i},\ldots,a_{d-1}) \in S_{+,j}$. The same holds for $S\setminus S_{+,i})$. Thus, the contribution of coordinates $a_i$ cancels in sums $\sum_{i>0}w_{i}a_{i}$ in Equation (\ref{eq:sym}), i.e.,

{
\begin{align} 
&-\nabla_{w_{0}} f(X,w) \nonumber\\
& =2\sum_{X \in S_{+,0}} (w^*_0-w_{0}) (1+d_0)\cdot(1+d_0) 
 +2\sum_{X \in S\setminus S_{+,0}}(w^*_0-w_{0}) (1-d_0)) \cdot(1-d_0) \nonumber \\
& =2|S_{+,0}| (w^*_0-w_{0}) (1+d_0)^2 
 +2|S\setminus S_{+,0}|(w^*_0-w_{0}) (1-d_0)^2 \nonumber \\ 
& =2|S| (w^*_0-w_{0}) (1+d_0^2) \label{eq:cow0}
\end{align} }%

Similarly for $w_{i>0}$:

{
\begin{align} 
-\nabla_{w_{i}} f(X,w) 
&= 2\sum_{X \in S_{+,i}}  \big((w^*_0-w_{0}) (1+a_0) - \sum_{j \neq i,j>0} w_{j}a_j-w_i(-d_i)\big)\cdot -d_i \nonumber \\
&+ 2\sum_{X \in S\setminus S_{+,i}}  \big((w^*_0-w_{0}) (1+a_0)) - \sum_{j \neq i,j>0} w_{j}a_j-w_i(d_i)\big) \cdot d_i \nonumber \\
\end{align}}%
As before the contribution of coordinates $j \neq i$ cancels for $i>0$. For the first coordinate $w_0$ it holds $(w^*_0-w_{0}) ((1+d_0)+(1-d_0)) = 2(w^*_0-w_{0})$.
{
\begin{align} 
-\nabla_{w_{i}} f(X,w) 
&= 2\sum_{X \in S_{+,i}}  \big(2(w^*_0-w_{0}) -w_i(-d_i)\big) \cdot(-d_i) \nonumber \\
&+ 2\sum_{X \in S\setminus S_{+,i}} \big(2(w^*_0-w_{0})  -w_i(d_i)\big) \cdot(d_i) \nonumber \\
&= -2|S|\big(2(w^*_0-w_{0})+ w_id_i\big)\cdot d_i \label{eq:cow1}
\end{align}}%
Plugging Equations (\ref{eq:cow0}) and (\ref{eq:cow1}) into Equation (\ref{eq:grad}) completes the proof.
\end{proof}

\paragraph{Explanation of Theorem \ref{thm:rec}:} From Equations (\ref{eq:rec1}) and (\ref{eq:rec2}), it becomes apparent that the magnitude of change of the coordinates $w_0$ and $w_{i}$ (for $i>0$) can differ strongly. For $i>0$, the factor of $d_i$ in Equation (\ref{eq:rec2}) implies that for small $d_{i}$, $w_i$ moves slowly towards $w^*_i=0$ (in the orthogonal direction of $w^*$). In contrast, even for small $d_0$, $w_0$ tends to move more quickly towards the optimal $w^*_0$. This follows from Equation (\ref{eq:rec1}) due to the term $(1+d_0^2)$: The change of $w_0$ is always at least proportional to the gap $w^*_0-w_0$ irrespective of $d_0$. Thus, dimensions, where the optimal vector $w^*$ is large, converge quickly, whereas dimensions, where the optimal vector $w^*$ is small, converge more slowly. Noisy dimensions (irrelevant dimensions, where the optimal vector is small) take a long time to be ``eliminated''. From the perspective of correlated initialization, a difference in length of the initial and the optimal vector is not a major concern since this gap can be fairly quickly reduced. This shows that it is preferable if the initial vector points in the same direction as the optimal.  Correlation of weights implies that weights are similar. Therefore, if an optimal weight vector has similar weights, initial weights should also have similar weights, i.e., weights should be correlated.
\paragraph{Loosening Assumptions of Theorem \ref{thm:rec}:}
The assumptions made for Theorem \ref{thm:rec} allow deriving intuitive, fairly simple closed-form expressions. We shall discuss the implications of loosening them in relation to Theorem \ref{thm:rec}.\\
\noindent\textbf{Loosening Assumption (\ref{ass:sym}):} The assumption says that samples are placed symmetrically on a grid, i.e., a coordinate $a_i$ can have only two values $a_i \in \{-d_i,d_i\}$. This helps in ensuring symmetry, i.e., technically they lead to cancellation of terms in Equation (\ref{eq:sym}), since $\sum_{X=(a_0,\ldots,a_{p-1}) \in S} a_i =0$ for all $i>0$. Breaking symmetry for a coordinate $a_i$ (by adding more points on one side or changing values of samples or both) might either increase or decrease convergence speed (for all other coordinates). If the average distance to the initial weight vector $w_0$ increases, it is increased. Compared to a symmetric distribution, more points are further away from $w_0$. Otherwise, it decreases. This follows from Equation (\ref{eq:sym}), where contributions of individual coordinates do not cancel, but residual terms capturing the asymmetry remain. Considering all coordinates, the impact of a non-symmetric distribution is likely small since some coordinates increase convergence speed, and others are expected to decrease it.

\noindent\textbf{Loosening Assumption (\ref{ass:opt}):} We assume that ground truth values $y$ can (in principle) be exactly reconstructed by the model, i.e., $y(X)=w^*X$. Adding noise or highly non-linear terms $f(X)$ to the outputs, i.e., assuming $y(X)=w^*X+f(X)$, can break locality. That is, two points $X, X'$ that are near each other ($|X-X'|$ small) might have a very different impact on the gradient. Technically, another term is added in Equation (\ref{eq:nab}) for $w\cdot X>0$, i.e., $-2f(X)\cdot X$. In general, the overall impact on convergence speed for a fixed initial condition $w_0$ cannot be anticipated without knowing $f$. 
For random noise, the overall impact might be small, since noise terms cancel.

\noindent\textbf{Loosening Assumption (\ref{ass:act}):} We assumed that initially all samples $X \in S$ are active, i.e., $w\cdot X>0$. Since $X_i$ might be said to be larger 0 because they are often the outcome of applying a $ReLU$ unit, the assumption boils down to assuming that all coordinates $w_i$ are non-negative throughout the optimization process. That is, in particular, the optimal solution $w^*$ has non-negative coordinates $w^*_i$. Setting the initial coordinate $w^{(0)}_0$ to a negative value might violate the assumption that input samples are active, i.e., $w\cdot X>0$. If for all $X$ holds $w\cdot X\leq0$, there are no updates (known as ``dead ReLU'') and $w$ does not converge to $w^*$. Let the number of active samples be $|S_A|$, i.e., samples $X \in S_A$  fulfill $w\cdot X>0$. Generally, the fewer active samples (the smaller $|S_A|$) the slower convergence is.  Let us derive this using Equation (\ref{eq:grad}):  
{
\begin{align}
w^{(i+1)}&=w^{(i)}-\frac{\lambda}{|S|} \sum_{X \in S}  \nabla_w f(X,w) \nonumber \\
&=w^{(i)}-\frac{\lambda}{|S|} (\sum_{X \in S_A}  \nabla_w f(X,w)+\sum_{X \in S\setminus S_A}  \nabla_w f(X,w)) \nonumber \\
&=w^{(i)}-\frac{\lambda}{|S|} (\sum_{X \in S_A}  \nabla_w f(X,w)+\sum_{X \in S\setminus S_A}  0)  \phantom{avb}\text{(Using Equation (\ref{eq:nab}))}\nonumber\\
&=w^{(i)}-\frac{\lambda}{|S|} \sum_{X \in S_A}  \nabla_w f(X,w) \label{eq:dead}
\end{align}}%

Therefore, for a concentrated set $S$ and $w$ being far away from the convex hull of $S_A$, it holds that updates to weights are roughly proportional to $|S_A|/|S|$ as seen in Equation \ref{eq:dead}. Thus, convergence is slow for few active samples $|S_A|$. In the case of many initially inactive samples, convergence might become faster and then slower again. It becomes faster since samples might eventually become active. It slows down when the vector is roughly of final length, but direction is not yet very well aligned.

\section{Multi Layer Variance} \label{sec:multi}
We investigate variance of outputs of the final layer $Var[y_l]$ for four scenarios (w/o perfect correlation of inputs, w/o perfect correlated initialization) shown in Table \ref{tab:var}. We show that in case both inputs and initialization are perfectly correlated, variance is significantly higher than for any other scenario, i.e., it is the squared value. Thus, our results indicate that if both inputs and initialization are correlated, variance of outputs increases. Prior work \cite{he15} made the assumption that inputs and initialization are uncorrelated, which led to the conclusion that weights should be initialized so that $Var[w_{i,j}]=2/k^2$, where $i$ denotes the layer index, $j$ the index of the weight and $k$ being the filter width (and height). In the light of our following derivation, the variance should be less, i.e., in between $[2/k^4,2/k^2]$. In practice, due to batch normalization and the fact that correlation of weights within a layer(see Figure \ref{fig:cor}) and initialization is typically far from perfectly correlated, the impact is small, and the benefits of adjusting the variance might be limited. Still, the insights are interesting from a theoretical perspective.

\begin{table}[h] 	
 	\begin{center}
 		\small
 		\setlength\tabcolsep{2.5pt}
 		\begin{tabular}{|l|l| l | l| }\hline
 		 \multicolumn{2}{|c|}{ } & \multicolumn{2}{|c|}{\footnotesize{Initialization}} \\ \cline{3-4}
 		 \multicolumn{2}{|c|}{ } & Perfect Corr. & Uncorrelated \\ \hline
 		 Inputs to & Perfect Corr. & (A) $\eta\cdot k^{2l}$ & (B) $\eta\cdot k^{l}$ \\ \cline{2-4}
 		 1st layer& Uncorrelated & (C) $\eta\cdot k^{l}$ &(D) $\eta\cdot k^{l}$ \cite{he15} \\ \cline{1-4}
 		\end{tabular}
 	\end{center}
 	\caption{Variance of layer $l$ depending on correlation of initialization and inputs to network for filter width $k$ and $\eta:=Var[y_{1,j}] (1/2\cdot Var[w_{0,j}])^l$} \label{tab:var}
\end{table}
We consider a similar model with similar analysis as in \cite{he15}. For ease of notation (avoiding another index), we consider a single spatial filter. That is, inputs have just one channel, and outputs consist of a single channel. A response of a convolutional layer is: $$y_{l}:= W_{l}x_{l}+b_l$$
where $x_{l}$ is a $k^2$-by-1 vector that represented co-located $k \times k$ pixels for filters of width $k$ and $c$ input channels. $W_{l}$ is a $1 \times k^2$ matrix. $b$ are the biases, which are initialized to 0, and the $y$ is the response at a pixel of the output map. $l$ denotes the layer. The input to layer $l$ is given by $x_l:=f(y_{l-1})$, where $f$ is the activation. We shall use ReLU $f(x)=\max(x,0)$. All weights $W_l$ have the same distribution with mean 0, being symmetric around the mean. Weights $W_l$ and inputs $x_l$ are independent. 
We start by investigating the scenario, where inputs to the network are perfectly correlated (per channel), i.e., $x_{0,j}=x_{0,j'}$ for any $j,j'$. Since the inputs $x_{0}$ to the first layer are perfectly correlated, also the inputs $x_{l}$ to any other layer are perfectly correlated, i.e., $x_{l,j}=x_{l,j'}$ $\forall l,j,j'$. Note, this holds irrespective of how weights are initialized. To see, this consider the second layer. It holds: $y_{1,j} = \sum_{t<d} \sum_{j'<k^2c} x_{0,j'}\cdot W_{0,j,j'} = x_{0,0} \sum_{j'<k^2} W_{0,j,j'}$. Therefore, the output $x_{1,j}=\max(y_{1,j},0)$ is the same irrespective of the position $j$.

We use the following fundamental laws:
{
\begin{align}
&Var(aX)=a^2X \label{var:const}\\
&Var(XY)=E[X^2]E[Y^2]-E[X]^2E[Y]^2 \text{\phantom{ab}(if $X,Y$ independent)} \label{var:ind}
\end{align}}%

We start with analyzing the variance of the final layer for correlated inputs and initialization (Case (A) in Table \ref{tab:var}):
{
\begin{align} 
Var[y_{i+1,j}]&=Var[\sum_{t<k^2} x_{i,j+t}\cdot w_{i,t}] =  Var[k^2x_{i,0}\cdot w_{i,0}] \nonumber \\ 
&=  k^4 Var[x_{i,0}\cdot w_{i,0}] \text{\phantom{ab}(Equation \ref{var:const})} \nonumber\\
&=k^4 E[(x_{i,0})^2]E[(w_{i,0})^2] \text{\phantom{ab}(Equation \ref{var:ind})} \label{eq:var}
\end{align}}%

Since $w_{i,j}$ has zero mean, it holds  $E[(w_{i,j})^2]=Var[w_{i,j}]$. Since $w_{i-1,j}$ is assumed to be symmetric around its mean, $x_{i-1,j}$ has zero mean and is symmetric around zero. This implies for $x_{i,j}$: $E[(x_{i,j})^2]=Var[y_{i-1,j}]/2$. Furthermore, since weights $w_{i,j}$ of all layers are distributed equally, it holds $Var[w_{i,j}]=Var[w_{0,j}]$:

{
\begin{align} 
Var[y_{i+1,j}]&=k^2 E[(x_{i,j})^2]E[(w_{i,j})^2] 
=k^4/2 Var[w_{i,j}]Var[y_{i,j}] \nonumber\\ 
Var[y_{l,j}]&=Var[y_{1,j}] (k^4/2 Var[w_{0,j}])^l \label{eq:yl}
\end{align}}%

Cases (B), (C), and (D) in Table \ref{tab:var}, where either inputs, weights or both are correlated are analyzed analogously. The key difference lies in the derivation of Equation \ref{eq:var}. For each case, at least one of two conditions $x_{i,0}=x_{i,j+t}$ and  $w_{i,t}=w_{i,0}$ is violated. In turn, we cannot use Equation \ref{var:const}. For Case (B)  we get:
{
\begin{align}
Var[y_{i+1,j}]&=Var[\sum_{t<k^2} x_{i,j+t}w_{i,t}]  =  Var[x_{i,j} \sum_{t<k^2} w_{i,t}] \nonumber\\
&=E[(x_{i,j})^2]E[(\sum_{t<k^2} w_{i,t})^2] \nonumber\\& \text{\phantom{ab} Using Equation (\ref{var:ind}) with } Y= \sum_{t<k^2} w_{i,t} \nonumber\\
&=E[(x_{i,j})^2](\sum_{t,t'<k^2} E[ w_{i,t}\cdot w_{i,t'}]) \nonumber\\
&=E[(x_{i,j})^2](\sum_{t<k^2} E[ (w_{i,t})^2] 
 + \sum_{t,t'<k^2, t\neq t'} E[ w_{i,t}]E[w_{i,t'}] ) \nonumber\\ 
&=k^2 E[(x_{i,j})^2]E[ (w_{i,t})^2] \text{\phantom{ab} since $E[ w_{i,t}]=0$} \label{eq:var2}
\end{align}}%
The only difference between Equation (\ref{eq:var}) and Equation (\ref{eq:var2}) is the factor $k$. Equation (\ref{eq:var2}) is used to derive an expression for $Var[y_{l,j}]$ with the same steps as the derivation of Equation (\ref{eq:yl}). 
{
\begin{align} 
Var[y_{i+1,j}]&=k^2 E[(x_{i,j})^2]E[(w_{i,j})^2] 
=k^2/2 Var[w_{i,j}]Var[y_{i,j}] \nonumber\\ 
Var[y_{l,j}]&=Var[y_{1,j}] (k^2/2Var[w_{0,j}])^l \label{eq:yl2}
\end{align}}%

Case (C) is analogous. Case (D) is identical to \cite{he15}.

\section{Correlated Initialization} \label{sec:corr}
We propose correlated initialization for a single spatial $k \times k$ filter. According to theory (Theorem \ref{thm:rec}) correlation of initialized weights should be similar to the final filter. Therefore, we estimate the correlation matrix $R$ of filters across layers of the entire network and use it to initialize filters.\footnote{Since correlation varies across layers (Figure \ref{fig:cor}) one might also employ a layer-wise estimation.} For computing the correlation matrix $R$ we utilize a trained network. We shall show that correlation matrices tend to be transferable across networks and datasets.\\ 
Algorithm \ref{alg:CorInit} illustrates the initialization for one spatial filter. These can be stacked to obtain the initialization of an entire layer. It uses standard statistical techniques to obtain samples using a correlation matrix. It requires the correlation matrix $R$ as input, which is computed based on weights of (single) filters as observations from an arbitrary trained network. The correlated (initialization) matrix is computed by decomposing the correlation matrix $R$ into $R=C\cdot C^T$ and then multiplying $C$ with the matrix $S_{unCorr}$ consisting of independently chosen random elements. Correlated initialization impacts variance.  The variance of layer outputs should be 1, so that variance remains constant across layers as commonly done, e.g. \cite{he15} and \cite{glo10}. We scale by $Var(w)$, computed under the assumption that weights are initialized using the Gaussian distribution $N(0,1)$ and the correlation matrix $R$. For a spatial  $k \times k$ filter $w$ holds:
{
\begin{align} 
Var(w) &=\sum_{i,j,l,m < k} Cov(w_{i,j},w_{l,m}) =\sum_{i,j<k^2} R(i,j) 
  =\sum_{i<k^2} R(i,i) + \sum_{i,j <k^2, i\neq j} R(i,j) \nonumber\\ 
  &=k^2 +\sum_{i,j <k^2, i\neq j} R(i,j) \label{eq:corrMat}
\end{align}}%
In Equation (\ref{eq:corrMat}) the first term $k^2$ is identical for uncorrelated initialization, i.e. it is just the sum of variances of individual weights $\sum_{i<k^2} R(i,i)=\sum_{i,j<k} Var(w_{i,j})$, while the second term is due to the fact that weights are not initialized independently.

\begin{algorithm}[htp]
	\caption{Correlated Init for Single Spatial Filter with He\cite{he15} scaling\\
	\textbf{Input:} KernelSize $k$, In-channels $n_{in}$, Correlation matrix $R$\\
	\textbf{Output:} Initialization for a $ k\cdot k$ filter}  \label{alg:CorInit}
	\begin{algorithmic}[1]
		\begin{small}			
		    \STATE Solve $C C^T = R$ to get $C$ (e.g. using Cholesky Decompostion)
		    \STATE $S_{unCorr}:=$ $k\times k$ matrix
		    initialized with independent random numbers from $N(0,1)$
		    \STATE $S_{corr}:=C\cdot S_{unCorr}$ \COMMENT{(Unscaled) Correlated Initialization}\\
		    \STATE $Var(S_{corr}):=\sum_{i,j<k} S_{corr}(i,j)$ \COMMENT{For \cite{he15},\cite{glo10} this is $k^2$}
		    \STATE $S_{sca}:=\frac{1}{\sqrt{Var(S_{corr}) n_{in} }}$ \COMMENT{Scaling using fan-In as for \cite{he15}}
		    \STATE return $S_{sca}$
		\end{small}
	\end{algorithmic}	
\end{algorithm}

\section{Experiments and Evaluation} \label{sec:eva}
Our experiments aim to test possible conclusions from our theory by empirically analyzing properties of networks such as convergence and generalization as well as the interaction of initialization with other relevant design options and hyper-parameters such as learning rate and l2-regularization. In particular, we investigate the impact of the ``zig-zag'' behavior of parameter updates, meaning that after an initial phase, where learning is fast, gradients start to change signs between iterations. Based on Theorem \ref{thm:rec} (for a single filter) we conjecture the following:\\
 (i): Generalization is better for correlated initialization. Bad initialization might lead to getting stuck in worse local minima than correlated initialization. With correlation initialization, the movement towards an optimum is more direct. The `zig-zag' phase poses a risk of getting stuck at a poor local optimum.\\
 (ii): Convergence is slower for uncorrelated initialization. Correlated initialization should converge to a steady-state before uncorrelated initialization since the `zig-zag' phase is shorter.\\
 (iii): Both (i) and (ii) underlie the assumption that `zig-zag' of weights is more profound for uncorrelated initialization than for correlated initialization. Formally, we measure the `zig-zag' by computing the percentage of sign changes of the gradient $\nabla_w f(X,w)$ of all parameters $w \in R^d$, between consecutive updates $i$ an $i+1$. The function $sign$  returns 1 for positive numbers, -1 for negative and 0 for the number 0. It is applied element-wise for each weight of the $d$ entries in $\nabla W$. Formally, we sum up: 
{
\begin{align}
Gradient Sign Changes:=\frac{\sum_{j=0}^{d-1}|sign(\nabla_{w^{(i)}_j} f(X,w^{(i)}))-sign(\nabla_{w^{(i+1)}_j} f(X,w^{(i+1)}))|}{2d} \nonumber
\end{align}}%
 (iv): Large learning rates diminish returns of correlated initialization. If learning rates are large, it might take just one step to reach the `zig-zag' behavior for (all) parameters. That is, gradients change signs commonly between updates of parameters. For large learning rates most parameters might zig-zag very early on, independent of initialization. Large learning rates are expected to make initialization less influential, i.e., reduce gains in generalization. Since poor local minima that might arise due to bad initialization (as conjectured in (i)) can be escaped more easily.\\
 (v): L2 regularization diminishes returns of correlated initialization. L2 regularization makes weights more uniform, i.e., it increases their correlation. Thus, it helps to push initially uncorrelated weights towards solutions that generalize better, i.e., ones where weights are correlated.\\
We also compare correlation matrices computed from various trained networks and vary the degree of correlation with other initialization methods. A comparison of the impact of increased network depth and width is also added, but it did not affect differences between the two initialization schemes strongly.

\begin{figure}
		\centering	
			\includegraphics[width=\linewidth, ]{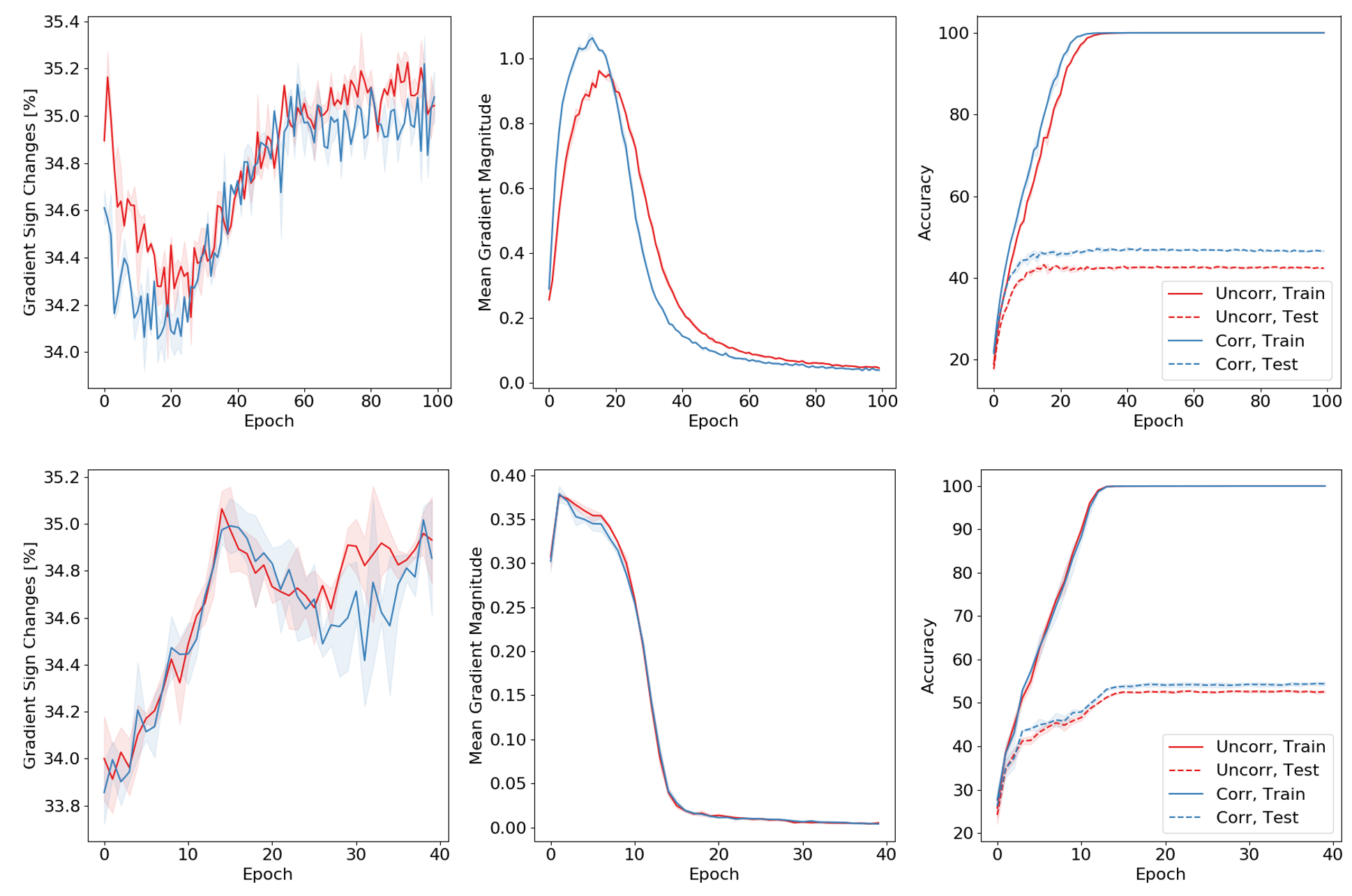}
		\caption{Gradient sign changes, gradient magnitude, and accuracy for VGG on CIFAR-100 using learning rate 0.005 (top row) and 0.1 (bottom row). The area within one standard deviation from mean is shaded in all plots. Correlated and uncorrelated initialization differ substantially showing behaviors aligned with theory. } \label{fig:convVGGCi100} 
\end{figure}

			

\subsection{Setup and Analysis} \label{sec:set}
We used a default setup that was varied in one or more aspects for individual experiments. Changes to the default setup are described for each experiment together with outcomes. SGD with momentum 0.8 with batch size 256 served for training without l2-regularization for 40 epochs. This was sufficient to achieve convergence  (see plots of accuracy in Figure \ref{fig:convVGGCi100}, \ref{fig:convVGGCi10}, \ref{fig:convResCi100}).  Since overfitting has been no significant concern, we report test performance for the fully trained network. We employed CIFAR-10\cite{kri09} without data augmentation and pre-defined split into training and test data. As networks, we used a VGG-8\cite{sim14} and a Resnet-10\cite{he16} and VGG-16 and Resnet-18 to compare with deeper networks  \footnote{Details of networks can be found in code under \url{https://drive.google.com/file/d/1lu93YTC8sNpMchiybCrA1rflw8uGvBKS/view?usp=sharing} since code could not be submitted and repository links are often not allowed due to anonymity policies; the default network size scaling for `netSize', i.e., neurons per layer, was 0.5 (and 1.5 for wide networks), batch-normalization `bn' was True.}.  We trained 5 networks for each configuration, i.e., hyperparameter setting. We report average accuracy and standard deviation (indicated as shaded area in plots). For initialization we used Kaiming normal\cite{he15} -- see Algorithm \ref{alg:CorInit}. For uncorrelated initialization, we used the identity matrix for $R$. Otherwise, we computed it from weights of a Resnet-10 network trained on CIFAR-10.

\subsection{Results}
\begin{figure}
		\centering	
			\includegraphics[width=\linewidth, ]{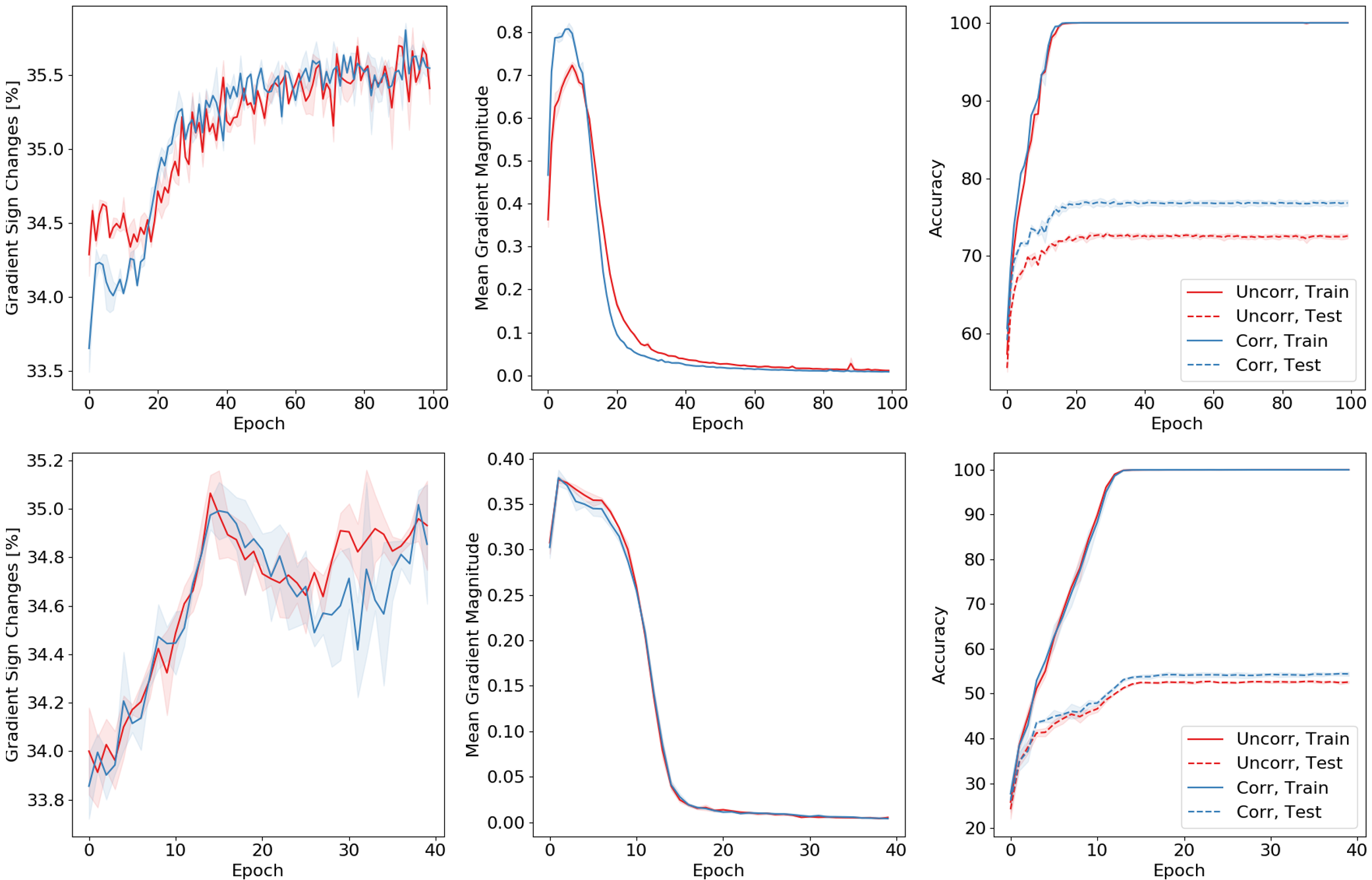}
		\caption{Gradient sign changes, gradient magnitude and accuracy for VGG on CIFAR-10 using learning rate 0.005 (top row) and 0.1 (bottom row). The area within one standard deviation from mean is shaded in all plots. Correlated and uncorrelated initialization differ substantially showing behaviors aligned with theory. } \label{fig:convVGGCi10} 
\end{figure}

\begin{figure}
		\centering	
			\includegraphics[width=\linewidth, ]{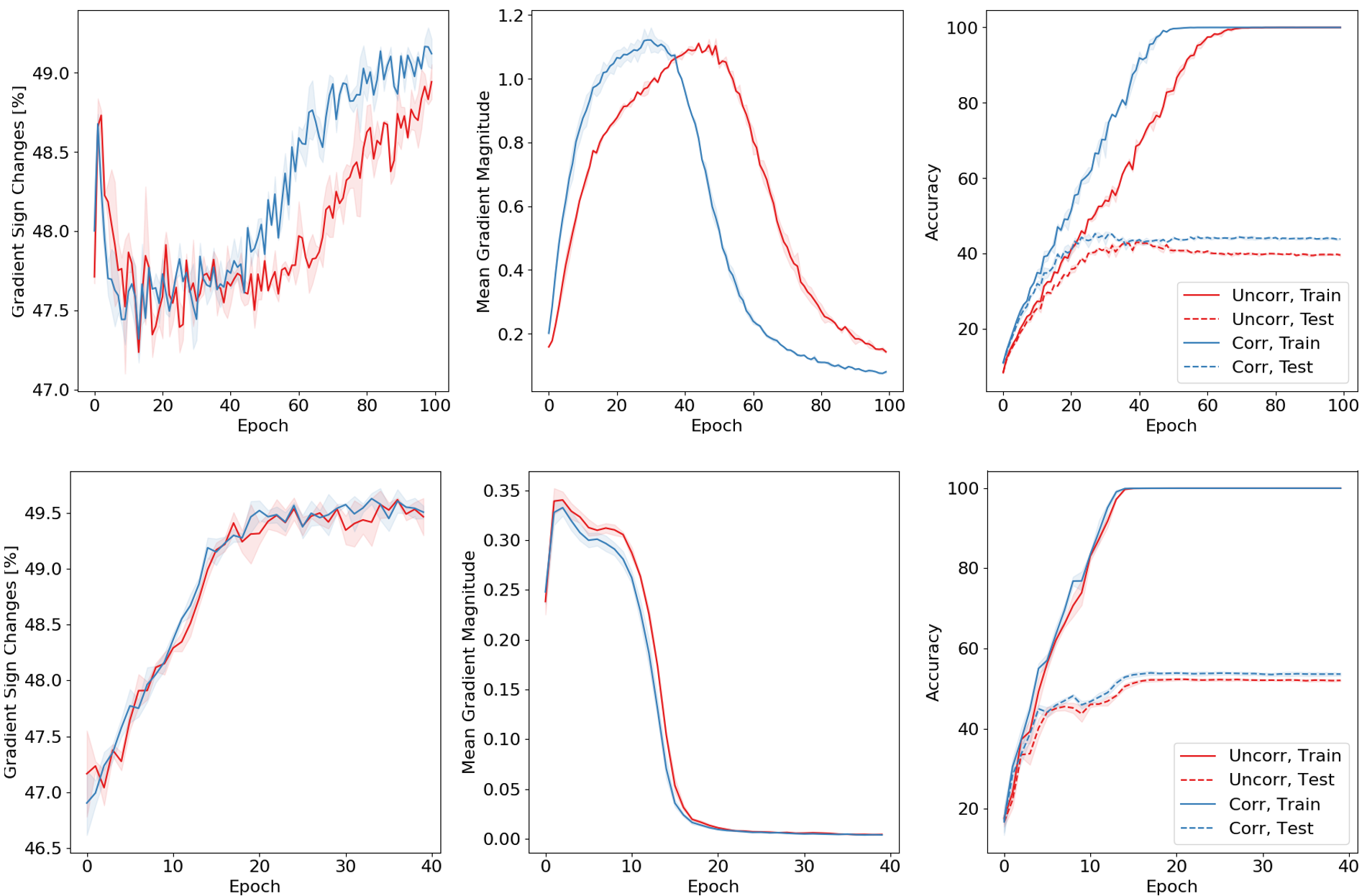}
		\caption{Gradient sign changes, gradient magnitude, and accuracy for ResNet on CIFAR-100 using learning rate 0.005 (top row) and 0.1 (bottom row). The area within one standard deviation from mean is shaded in all plots. Correlated and uncorrelated initialization differ substantially showing behaviors aligned with theory. } \label{fig:convResCi100} 
\end{figure}

A comparison of uncorrelated and correlated initialization for accuracy, gradient sign changes and gradient magnitudes can be seen for VGG on CIFAR-10 and CIFAR-100 using a low learning rate of 0.005 and a high learning rate of 0.1 in Figures \ref{fig:convVGGCi100} and \ref{fig:convVGGCi10} and for Resnet-10 for CIFAR-100 in Figure \ref{fig:convResCi100} (CIFAR-10 behaves similarly for ResNet. We show quantitative results for CIFAR-10 and ResNet).

\smallskip

\noindent\textbf{Uncorrelated and correlated initialization for low learning rate:} 
For VGG, the rightmost panel in the top row of Figures \ref{fig:convVGGCi10} and \ref{fig:convVGGCi100} shows that generalization is significantly better for correlated initialization for a low learning rate, hinting that the initial solution is closer to a good local optimum that the parameters converge to.  Training accuracy is also constantly higher. The early gap in accuracy is essentially maintained throughout training. Furthermore, the average gradient magnitude is larger (central panel) for correlated initialization in the first few epochs. At the same time, the zig-zagging, i.e., sign changes between two continuous updates, is lower. Thus, uncorrelated initialization leads to smaller, seemingly more random updates lacking a consistent direction. 
For Resnet-10 the qualitative behavior is similar though less profound, as shown in Figure \ref{fig:convResCi100} in the upper plots for low learning rate. In particular, differences in gradient sign changes in the initial phase can only be seen upon very careful investigation. Smaller differences are not unexpected since for Resnets we compute for each block $y(X)=F(X)+X$. The addition of $X$ naturally encourages correlation between inputs and outputs. Note that $y(X)=X$ leads to a correlation of 1. Therefore, given that for $y(X)=F(X)$ we have that $y(X)$ and $F(X)$ are not or only weakly correlated, the addition of $X$ increases correlation.

\smallskip

\noindent \textbf{Impact of learning rate:} For a low learning rate (top rows in Figures \ref{fig:convVGGCi100}, \ref{fig:convVGGCi10} and \ref{fig:convResCi100}) differences between both initialization schemes are more pronounced than for high learning rates (bottom rows in Figures \ref{fig:convVGGCi100}, \ref{fig:convVGGCi10} and \ref{fig:convResCi100}). This is expected since for high learning rates also correlated initialization quickly ends up in a zig-zagging phase. Visually speaking, in the right panel of Figure \ref{fig:wmul} less iterations are needed until the weight vector ends up moving mainly back and forth and only slowly towards the optimal solution. 
However, differences in generalization behavior remain, i.e. correlated initialization still outperforms. Numbers are shown for both datasets and networks for high learning rates in Table \ref{tab:l2}. 


\begin{table}
	\centering
	\scriptsize
\begin{tabular}{|c| c|c|c| c|c|c| } \hline 
$L2$ Regularization  & \multicolumn{2}{|c|}{ ResNet-10}&  \multicolumn{2}{|c|}{VGG-8}\\ \cline{2-5}
Parameter $\lambda$ & CIFAR-10&CIFAR-100& CIFAR-10&CIFAR-100\\ \hline
 0 (Correlated) & 0.826 \tiny{$\pm$ 0.002} & 0.536 \tiny{$\pm$ 0.004} & 0.827 \tiny{$\pm$ 0.002} & 0.544 \tiny{$\pm$ 0.004} \\ \hline
0 (Uncorrelated)  & 0.815 \tiny{$\pm$ 0.002} & 0.52 \tiny{$\pm$ 0.002} & 0.815 \tiny{$\pm$ 0.003} & 0.526 \tiny{$\pm$ 0.002} \\ \hline
0.001 (Correlated) & 0.849 \tiny{$\pm$ 0.001} & 0.554 \tiny{$\pm$ 0.004} & 0.834 \tiny{$\pm$ 0.003} & 0.551 \tiny{$\pm$ 0.004} \\ \hline
0.001 (Uncorrelated)  & 0.851 \tiny{$\pm$ 0.002} & 0.548 \tiny{$\pm$ 0.002} & 0.832 \tiny{$\pm$ 0.002} & 0.545 \tiny{$\pm$ 0.004} \\ \hline
\end{tabular}
	\caption{Impact of L2-regularization. L2-regularization is more beneficial for uncorrelated initialization.}\label{tab:l2}
\end{table}

\smallskip

\noindent\textbf{L2-regularization:} \label{sec:ove}
Adding L2 regularization leads to slower convergence; therefore, we increased the number of epochs to 80 for positive regularization parameter $\lambda=0.001$ to ensure convergence. Table \ref{tab:l2} shows that adding L2-regularization leads to gains for uncorrelated and correlated initialization, but as expected, uncorrelated initialization benefits more.

\begin{table}
	\centering
	\scriptsize
	\begin{tabular}{|c| c|c| c|c| } \hline 
Initialization& \multicolumn{2}{|c|}{ ResNet-10}&  \multicolumn{2}{|c|}{VGG-8}\\ \cline{2-5}
Method & CIFAR-10&CIFAR-100& CIFAR-10&CIFAR-100\\ \hline
kaiming-normal (Correlated) & 0.826 \tiny{$\pm$ 0.002} & 0.536 \tiny{$\pm$ 0.004} & 0.827 \tiny{$\pm$ 0.002} & 0.544 \tiny{$\pm$ 0.004} \\ \hline
kaiming-normal (Uncorrelated)  & 0.815 \tiny{$\pm$ 0.002} & 0.52 \tiny{$\pm$ 0.002} & 0.815 \tiny{$\pm$ 0.003} & 0.526 \tiny{$\pm$ 0.002} \\ \hline
orthogonal & 0.814 \tiny{$\pm$ 0.003} & 0.521 \tiny{$\pm$ 0.005} & 0.816 \tiny{$\pm$ 0.003} & 0.531 \tiny{$\pm$ 0.004} \\ \hline
xavier-normal & 0.811 \tiny{$\pm$ 0.002} & 0.521 \tiny{$\pm$ 0.006} & 0.817 \tiny{$\pm$ 0.003} & 0.533 \tiny{$\pm$ 0.002} \\ \hline
\end{tabular}
	\caption{Comparison of Initialization Methods}\label{tab:init}
\end{table}

\smallskip

\noindent\textbf{Comparing Initialization Methods:} Table \ref{tab:init} shows a comparison with other common initialization methods. Correlated initialization performs best for both datasets and architectures. All uncorrelated initialization methods performed similarly well except for CIFAR-100 and VGG-8, where Kaiming-normal with uncorrelated initialization performed clearly worst.
\begin{table}
	\centering
	\scriptsize
\begin{tabular}{|c| c|c| c|c| } \hline 
Source for & \multicolumn{2}{|c|}{ ResNet-10}&  \multicolumn{2}{|c|}{VGG-8}\\ \cline{2-5}
correlation matrix $R$ & CIFAR-10&CIFAR-100& CIFAR-10&CIFAR-100\\ \hline
VGG-8 trained on CIFAR-10 & 0.814 \tiny{$\pm$ 0.002} & 0.527 \tiny{$\pm$ 0.005} & 0.818 \tiny{$\pm$ 0.002} & 0.535 \tiny{$\pm$ 0.002}  \\ \hline
Resnet-10 trained on CIFAR-10 & 0.826 \tiny{$\pm$ 0.002} & 0.536 \tiny{$\pm$ 0.004} & 0.827 \tiny{$\pm$ 0.002} & 0.544 \tiny{$\pm$ 0.004} \\ \hline 
None (Uncorrelated) & 0.815 \tiny{$\pm$ 0.002} & 0.52 \tiny{$\pm$ 0.002} & 0.815 \tiny{$\pm$ 0.003} & 0.526 \tiny{$\pm$ 0.002} \\ \hline
\end{tabular}
	\caption{Comparison of impact of source for correlation matrix $R$ in Algo. \ref{alg:CorInit}}\label{tab:corrsou}
\end{table}

\smallskip

\noindent \textbf{Comparing Correlation Matrices:} \label{sec:coma} Table \ref{tab:corrsou} shows a comparison with correlation matrices estimated from either VGG trained on CIFAR-10 or Resnet trained on CIFAR-10. Correlated initialization based on Resnet performs best for both datasets and architectures. Initialization based on VGG differences are marginal for CIFAR-10 and only significant for CIFAR-100 compared to uncorrelated initialization. We found that the correlations for Resnet are overall significantly stronger. The correlation between two locations $(i,j),(k,l)$ in a spatial filter are 0.26 for Resnet and only 0.18 for VGG. Furthermore, VGG also exhibits more and slightly stronger negative correlations, e.g., 50\% vs 20\% and 0.04 vs. 0.05. This suggests that increasing correlations beyond the level found in trained networks might be beneficial. Also, we conjecture that Resnets exhibit better localization than VGG, e.g., location information of features is more precise due to the use of residual layers that skip layers.

\begin{table}
	\centering
	\scriptsize
\begin{tabular}{|c| c|c| c|c| } \hline 
Strength & \multicolumn{2}{|c|}{ ResNet-10}&  \multicolumn{2}{|c|}{VGG-8}\\ \cline{2-5}
Factor $\alpha$ & CIFAR-10&CIFAR-100& CIFAR-10&CIFAR-100\\ \hline
1.5 & 0.822\tiny{$\pm$0.004} & 0.531\tiny{$\pm$0.005} & 0.821\tiny{$\pm$0.004} & 0.537\tiny{$\pm$0.004}  \\ \hline
1.0 &  0.82\tiny{$\pm$0.004} & 0.529\tiny{$\pm$0.006} & 0.821\tiny{$\pm$0.004} & 0.535\tiny{$\pm$0.004} \\ \hline 
0.5 & 0.817\tiny{$\pm$0.004} & 0.527\tiny{$\pm$0.005} & 0.819\tiny{$\pm$0.003} & 0.533\tiny{$\pm$0.004} \\ \hline
0.25 & 0.816\tiny{$\pm$0.004} & 0.525\tiny{$\pm$0.005} & 0.817\tiny{$\pm$0.003} & 0.53\tiny{$\pm$0.004} \\ \hline

\end{tabular}
	\caption{Impact of correlation strength, i.e. modifying off diagonal entries in matrix $R$ by factor $\alpha$}\label{tab:corrstr}
\end{table}

\smallskip

\noindent\textbf{Comparing Correlation Strength:} Table \ref{tab:corrstr} shows a comparison with correlation matrices of different strengths. We used the base correlation matrix $R$ and multiplied all entries by a factor $\alpha$ (except diagonal entries that remain 1, i.e., they indicate correlation of entry with itself). The table indicates that a stronger correlation leads to better results.

\smallskip

\noindent\textbf{Impact of network width and depth:}  Table \ref{tab:depth} shows a comparison for deeper networks, i.e., ResNet-18 and VGG-16, and wider networks. Increasing depth and width improves performance for both initialization types. However, the gap between correlated and uncorrelated initialization remains.  Table \ref{tab:part} shows a comparison for even deeper networks, i.e., ResNet-34 and VGG-19 with and without L2-regularization and reduced correlation. We found that networks with correlated initialization still outperform though differences can be small for such deep networks. Furthermore, it can be advantageous to use weaker correlation.

\begin{table}
	\centering
	\footnotesize
\begin{tabular}{|c| l|l|l| l| } \hline 
  & \multicolumn{2}{|c|}{ ResNet-10}&  \multicolumn{2}{|c|}{VGG-8}\\ \cline{2-5}
 & CIFAR-10&CIFAR-100& CIFAR-10&CIFAR-100\\ \hline
 Correlated Init& 0.826 \tiny{$\pm$ 0.002} & 0.536 \tiny{$\pm$ 0.004} & 0.827 \tiny{$\pm$ 0.002} & 0.544 \tiny{$\pm$ 0.004} \\ \hline
  Uncorrelated& 0.815 \tiny{$\pm$ 0.002} & 0.52 \tiny{$\pm$ 0.002} & 0.815 \tiny{$\pm$ 0.003} & 0.526 \tiny{$\pm$ 0.002} \\ \hline\hline
   \multicolumn{5}{|l|}{Number of neurons per layer multiplied by 3 }\\ \hline
  & \multicolumn{2}{|c|}{ ResNet-10} &  \multicolumn{2}{|c|}{VGG-8}\\ \cline{2-5}
 & CIFAR-10&CIFAR-100& CIFAR-10&CIFAR-100\\ \hline
 Correlated Init& 0.85 \tiny{$\pm$ 0.001} & 0.602 \tiny{$\pm$ 0.001} & 0.846 \tiny{$\pm$ 0.001} & 0.622 \tiny{$\pm$ 0.003} \\ \hline
  Uncorrelated& 0.837 \tiny{$\pm$ 0.002} & 0.579 \tiny{$\pm$ 0.0} & 0.84 \tiny{$\pm$ 0.001} & 0.606 \tiny{$\pm$ 0.001} \\ \hline\hline
  
  \multicolumn{5}{|l|}{Deeper networks }\\ \hline
  & \multicolumn{2}{|c|}{ ResNet-18}&  \multicolumn{2}{|c|}{VGG-16}\\ \cline{2-5}
 & CIFAR-10&CIFAR-100& CIFAR-10&CIFAR-100\\ \hline
 Correlated Init& 0.849 \tiny{$\pm$ 0.002} & 0.575 \tiny{$\pm$ 0.004} & 0.856 \tiny{$\pm$ 0.002} & 0.571 \tiny{$\pm$ 0.004} \\ \hline
  Uncorrelated& 0.833 \tiny{$\pm$ 0.003} & 0.545 \tiny{$\pm$ 0.002} & 0.844 \tiny{$\pm$ 0.003} & 0.559 \tiny{$\pm$ 0.002} \\ \hline
  \end{tabular}
	\caption{Impact of network width and depth. Differences of correlated initialization are not strongly impacted -- in particular compared to other hyper-parameters such as learning rate and L2-regularization}\label{tab:depth}
\end{table}

 \begin{table}
	\centering
	\footnotesize
\begin{tabular}{|c| l|l|l| l| } \hline 
  \multicolumn{5}{|l|}{No L2-regularization: $\lambda=0$ }\\ \hline
  & \multicolumn{2}{|c|}{ ResNet-34}&  \multicolumn{2}{|c|}{VGG-19}\\ \cline{2-5}
 & CIFAR-10&CIFAR-100& CIFAR-10&CIFAR-100\\ \hline
 Correlated Init& 0.828\tiny{$\pm$0.012 } & 0.581\tiny{$\pm$0.008 } & 0.867\tiny{$\pm$0.005} & 0.567\tiny{$\pm$0.009 } \\ \hline
 
 Correlated Init $\alpha=0.25$ & 0.835\tiny{$\pm$0.014} & 0.573\tiny{$\pm$0.012 } & 0.869\tiny{$\pm$0.004  } & 0.568\tiny{$\pm$0.007} \\ \hline

 Uncorrelated& 0.827\tiny{$\pm$0.014 } & 0.566\tiny{$\pm$0.008 } & 0.867\tiny{$\pm$0.004 } & 0.568\tiny{$\pm$0.007} \\ \hline \hline
 
 \multicolumn{5}{|l|}{L2-regularization: $\lambda=0.001$ }\\ \hline
 & \multicolumn{2}{|c|}{ ResNet-34}&  \multicolumn{2}{|c|}{VGG-19}\\ \cline{2-5}
 & CIFAR-10&CIFAR-100& CIFAR-10&CIFAR-100\\ \hline
 Correlated Init& 0.877\tiny{$\pm$0.004 } & 0.592\tiny{$\pm$0.014 } & 0.896\tiny{$\pm$0.003  } & 0.639\tiny{$\pm$0.004  } \\ \hline
 
 Correlated Init $\alpha=0.25$ & 0.878\tiny{$\pm$0.004} & 0.601\tiny{$\pm$0.007 } & 0.896\tiny{$\pm$0.003  } & 0.634\tiny{$\pm$0.006} \\ \hline
 
 Uncorrelated & 0.874\tiny{$\pm$0.006  } & 0.599\tiny{$\pm$0.007      } & 0.894\tiny{$\pm$0.002}& 0.626\tiny{$\pm$0.005} \\ \hline
\end{tabular}
	\caption{Deep networks w/o L2-regularization and correlation weakened by factor $\alpha$. Weak correlation always matches or exceeds performance of uncorrelated networks.}\label{tab:part}
\end{table}

\section{Discussion} \label{sec:dis}
Deep learning is in need of theory to make sense of its behavior. We believe that ad-hoc methods from XAI\cite{mesk20,sch19,sch20exp} are helpful. For example, it might be interesting to understand through visualization what information is retained in the network\cite{sch20exp} depending on the initialization. But only works that aim to understand models and their training on a deeper, theoretical level can mitigate this issue. Our work is one of the rather few works investigating training dynamics of networks focusing on single works rather than on a statistical level, e.g., as in \cite{sax19}. Thus, it might inspire future works in this direction through designing and analyzing even more realistic scenarios, and models are difficult. Initialization is highly important though techniques such as batch-normalization and for correlated initialization, L2-regularization diminish their relevance or, put differently, compensate for poor initialization. 

\section{Conclusions} \label{sec:conc}
This work used theoretical and empirical analysis to deepen the understanding of networks leveraging correlated data. Our theory is one of the few that formalizes the learning dynamics of single weights. For a single spatial filter, we show that correlated initialization leads to fewer updates of parameters that are undone in the next update ('zig-zag'). Theory also shows that it can lead to better generalization. Across layers, correlated initialization leads to larger variance in updates that must be accounted for by scaling weights. Our empirical analysis confirms that correlated initialization can lead to faster training and better generalization. The impact of correlated initialization is diminished for large learning rates and L2-regularization.

\bibliographystyle{spmpsci} 
\bibliography{refs}

\begin{thebibliography}{10}
\providecommand{\url}[1]{{#1}}
\providecommand{\urlprefix}{URL }
\expandafter\ifx\csname urlstyle\endcsname\relax
  \providecommand{\doi}[1]{DOI~\discretionary{}{}{}#1}\else
  \providecommand{\doi}{DOI~\discretionary{}{}{}\begingroup
  \urlstyle{rm}\Url}\fi

\bibitem{adv20}
Advani, M.S., Saxe, A.M., Sompolinsky, H.: High-dimensional dynamics of
  generalization error in neural networks.
\newblock Neural Networks \textbf{132}, 428--446 (2020)

\bibitem{agu19}
Aguirre, D., Fuentes, O.: Improving weight initialization of relu and output
  layers.
\newblock In: International Conference on Artificial Neural Networks, pp.
  170--184 (2019)

\bibitem{arp19}
Arpit, D., Bengio, Y.: The benefits of over-parameterization at initialization
  in deep relu networks.
\newblock arXiv preprint arXiv:1901.03611  (2019)

\bibitem{bal17}
Balduzzi, D., Frean, M., Leary, L., Lewis, J., Ma, K.W.D., McWilliams, B.: The
  shattered gradients problem: If resnets are the answer, then what is the
  question?
\newblock In: Proc. of the International Conference on Machine Learning (ICML)
  (2017)

\bibitem{chen20}
Chen, H., Zheng, L., Al~Kontar, R., Raskutti, G.: Stochastic gradient descent
  in correlated settings: A study on gaussian processes.
\newblock Advances in neural information processing systems  (2020)

\bibitem{coa11}
Coates, A., Ng, A., Lee, H.: An analysis of single-layer networks in
  unsupervised feature learning.
\newblock In: Int. conf. on artificial intelligence and statistics(AIStats)
  (2011)

\bibitem{glo10}
Glorot, X., Bengio, Y.: Understanding the difficulty of training deep
  feedforward neural networks.
\newblock In: Int. conf. on artificial intelligence and statistics (2010)

\bibitem{gro19}
Grosse, K., Trost, T.A., Mosbach, M., Backes, M., Klakow, D.: Adversarial
  initialization--when your network performs the way i want.
\newblock arXiv preprint arXiv:1902.03020  (2019)

\bibitem{han18}
Hanin, B., Rolnick, D.: How to start training: The effect of initialization and
  architecture.
\newblock In: Advances in Neural Information Processing Systems (2018)

\bibitem{he19}
He, K., Girshick, R., Doll{\'a}r, P.: Rethinking imagenet pre-training.
\newblock In: Proc. of the International Conference on Computer Vision, pp.
  4918--4927 (2019)

\bibitem{he15}
He, K., Zhang, X., Ren, S., Sun, J.: Delving deep into rectifiers: Surpassing
  human-level performance on imagenet classification.
\newblock In: Proc. of the international conference on computer vision, pp.
  1026--1034 (2015)

\bibitem{he16}
He, K., Zhang, X., Ren, S., Sun, J.: Deep residual learning for image
  recognition.
\newblock In: Conference on computer vision and pattern recognition (CVPR), pp.
  770--778 (2016)

\bibitem{iof15}
Ioffe, S., Szegedy, C.: Batch normalization: Accelerating deep network training
  by reducing internal covariate shift.
\newblock arXiv preprint arXiv:1502.03167  (2015)

\bibitem{kotur17}
Koturwar, S., Merchant, S.: Weight initialization of deep neural networks
  (dnns) using data statistics.
\newblock arXiv preprint arXiv:1710.10570  (2017)

\bibitem{kri09}
Krizhevsky, A., Hinton, G.: Learning multiple layers of features from tiny
  images.
\newblock Tech. rep. (2009)

\bibitem{kum17w}
Kumar, S.K.: On weight initialization in deep neural networks.
\newblock arXiv preprint arXiv:1704.08863  (2017)

\bibitem{luo18}
Luo, P., Wang, X., Shao, W., Peng, Z.: Towards understanding regularization in
  batch normalization.
\newblock arXiv preprint arXiv:1809.00846  (2018)

\bibitem{meh20}
Mehta, H., Cutkosky, A., Neyshabur, B.: Extreme memorization via scale of
  initialization.
\newblock arXiv preprint arXiv:2008.13363  (2020)

\bibitem{mesk20}
Meske, C., Bunde, E., Schneider, J., Gersch, M.: Explainable artificial
  intelligence: objectives, stakeholders, and future research opportunities.
\newblock Information Systems Management pp. 1--11 (2021)

\bibitem{mis15}
Mishkin, D., Matas, J.: All you need is a good init.
\newblock arXiv:1511.06422  (2015)

\bibitem{san18}
Santurkar, S., Tsipras, D., Ilyas, A., Madry, A.: How does batch normalization
  help optimization?
\newblock In: Adv. in Neural Information Processing Systems (2018)

\bibitem{sax19}
Saxe, A.M., Bansal, Y., Dapello, J., Advani, M., Kolchinsky, A., Tracey, B.D.,
  Cox, D.D.: On the information bottleneck theory of deep learning.
\newblock Journal of Statistical Mechanics: Theory and Experiment
  \textbf{2019}(12), 124020 (2019)

\bibitem{sax13}
Saxe, A.M., McClelland, J.L., Ganguli, S.: Exact solutions to the nonlinear
  dynamics of learning in deep linear neural networks.
\newblock arXiv:1312.6120  (2013)

\bibitem{sch19loc}
Schneider, J.: Locality-promoting representation learning.
\newblock In: International Conference on Pattern Recognition (ICPR) (2021)

\bibitem{sch19}
Schneider, J., Handali, J.: Personalized explanation in machine learning: A
  conceptualization.
\newblock In: European Conference on Information Systems (2019)

\bibitem{sch20exp}
Schneider, J., Vlachos, M.: Explaining neural networks by decoding layer
  activations.
\newblock Intelligent Data Analysis  (2021)

\bibitem{sim14}
Simonyan, K., Zisserman, A.: Very deep convolutional networks for large-scale
  image recognition.
\newblock Int. Conference on Learning Representations (ICLR)  (2014)

\bibitem{yan17}
Yang, G., Schoenholz, S.: Mean field residual networks: On the edge of chaos.
\newblock In: Advances in neural information processing systems, pp. 7103--7114
  (2017)

\bibitem{zha19}
Zhang, H., Dauphin, Y.N., Ma, T.: Fixup initialization: Residual learning
  without normalization.
\newblock arXiv preprint arXiv:1901.09321  (2019)

\end{thebibliography}

\end{document}